\documentclass[10pt,twocolumn,letterpaper]{article}

\usepackage{btas}
\usepackage{times}
\usepackage{epsfig}
\usepackage{graphicx}
\usepackage{amsmath}
\usepackage{amssymb}

\btasfinalcopy 

\ifbtasfinal\fi
\begin{document}

\title{Very Low-Resolution Iris Recognition Via\\ Eigen-Patch
Super-Resolution and Matcher Fusion}

\author{Fernando Alonso-Fernandez\\
IS-Lab/CAISR\\
Halmstad University (Sweden)\\
{\tt\small feralo@hh.se}
\and
Reuben A. Farrugia\\
Department of CCE\\
University of Malta (Malta)\\
{\tt\small reuben.farrugia@um.edu.mt} \and
Josef Bigun\\
IS-Lab/CAISR\\
Halmstad University (Sweden)\\
{\tt\small josef.bigun@hh.se} }

\maketitle
\thispagestyle{empty}

\begin{abstract}
Current research in iris recognition is moving towards enabling more
relaxed acquisition conditions. This has effects on the quality of
acquired images, with low resolution being a predominant issue.
Here, we evaluate a super-resolution algorithm used to reconstruct
iris images based on Eigen-transformation of local image patches.
Each patch is reconstructed separately, allowing better quality of
enhanced images by preserving local information.
Contrast enhancement is used to improve the reconstruction quality,
while matcher fusion has been adopted to improve iris recognition
performance.
We validate the system using a database of 1,872 near-infrared iris
images. The presented approach is superior to bilinear or bicubic
interpolation, especially at lower resolutions, and the fusion of
the two systems pushes the EER to below 5\% for down-sampling
factors up to a image size of only 13$\times$13.
\end{abstract}

\section{Introduction}

Iris is regarded as the most reliable biometric modality
\cite{[Jain08handbookbiometrics]}. Current research trends move
towards more relaxed acquisition conditions, allowing `at a
distance' and `on the move' capabilities \cite{[Jain10]}.
However, this has implications in terms of quality of acquired
images, with the lack of pixel resolution being one of the most
evident.
This paper is concerned with the task of up-sampling, or increasing
both size and quality of a low-resolution image as a result, for
example, of long acquisition distances.
Another piece of research, not addressed here, is concerned with
compressed images but with image dimensions kept constant
\cite{[Quinn14]}.

\begin{figure}[htb]
     \centering
     \includegraphics[width=.3\textwidth]{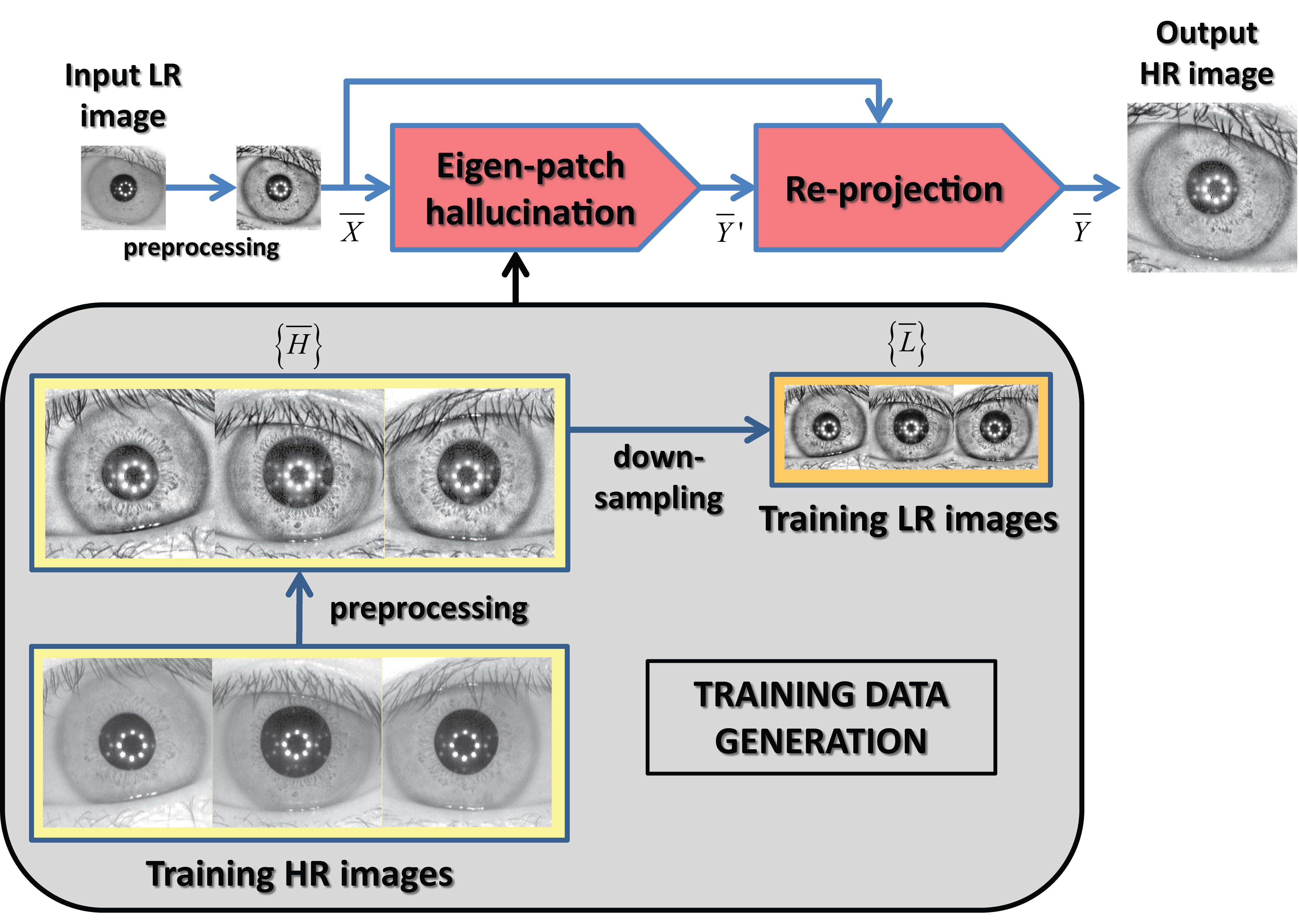}
     \caption{Eigen-patch iris hallucination system.}
     \label{fig:system}
\end{figure}

The aim of super-resolution (SR) techniques is to reconstruct the
missing high resolution (HR) image $\overline Y$ given a low
resolution (LR) image $\overline X$. The LR image is modeled as the
corresponding HR image manipulated by blurring $B$, warping $W$ and
down-sampling $D$ (plus some additive noise $\overline n$) as
$\overline X  = DBW\overline Y + \overline n$. For simplicity, some
works omit the warp matrix and noise, leading to $\overline X =
DB\overline Y$. Super-resolution can enhance the quality of LR
images, and thus the recognition performance. However, SR in
biometrics is relatively recent, with a lot of research in face
since 2002 (also called hallucination) \cite{[Wang14]}, and a much
more limited amount for iris and ocular modalities.
Despite the vast literature on image SR, one reason of such limited
research might be that most SR approaches are general-scene,
designed to produce overall visual enhancement, but the aim of
biometrics is a better recognition accuracy \cite{[Nguyen12]}.
Therefore, adaptation of super-resolution techniques to the
particularities of images from a specific biometric modality is
needed to achieve a more efficient up-sampling \cite{[Baker02]}.
Two main SR approaches exist: reconstruction- and learning-based
\cite{[Park03]}. In reconstruction-based, multiple LR images are
fused to obtain a HR image, therefore several LR images are needed
as input (which might not always be the case).
Alternatively, learning-based methods model the relationship between
LR and HR images with a training dictionary, using the learned model
to up-sample unseen LR images
Learning-based methods only need one LR image as input, and
generally outperform reconstruction-based methods, achieving higher
magnification factors \cite{[Park03]}.
%
%
The few works available on iris learning-based approaches employ
Multi-Layer Perceptrons \cite{[Shin09]}, or frequency analysis
\cite{[Deshpande14]}.
One major limitation is that they try to develop a prototype iris
using combination of complete images. Eigen-patches is a strategy
which models a local patch using collocated patches from a
dictionary, instead of using the whole image. Each patch is
hallucinated separately, providing better quality reconstructed
prototypes with better local detail and lower distortions. Local
methods are also generally superior in recovering texture than
global methods, which is essential due to the prevalence of
texture-based methods in ocular biometrics \cite{[Nigam15]}.

\begin{figure}[htb]
     \centering
     \includegraphics[width=.35\textwidth]{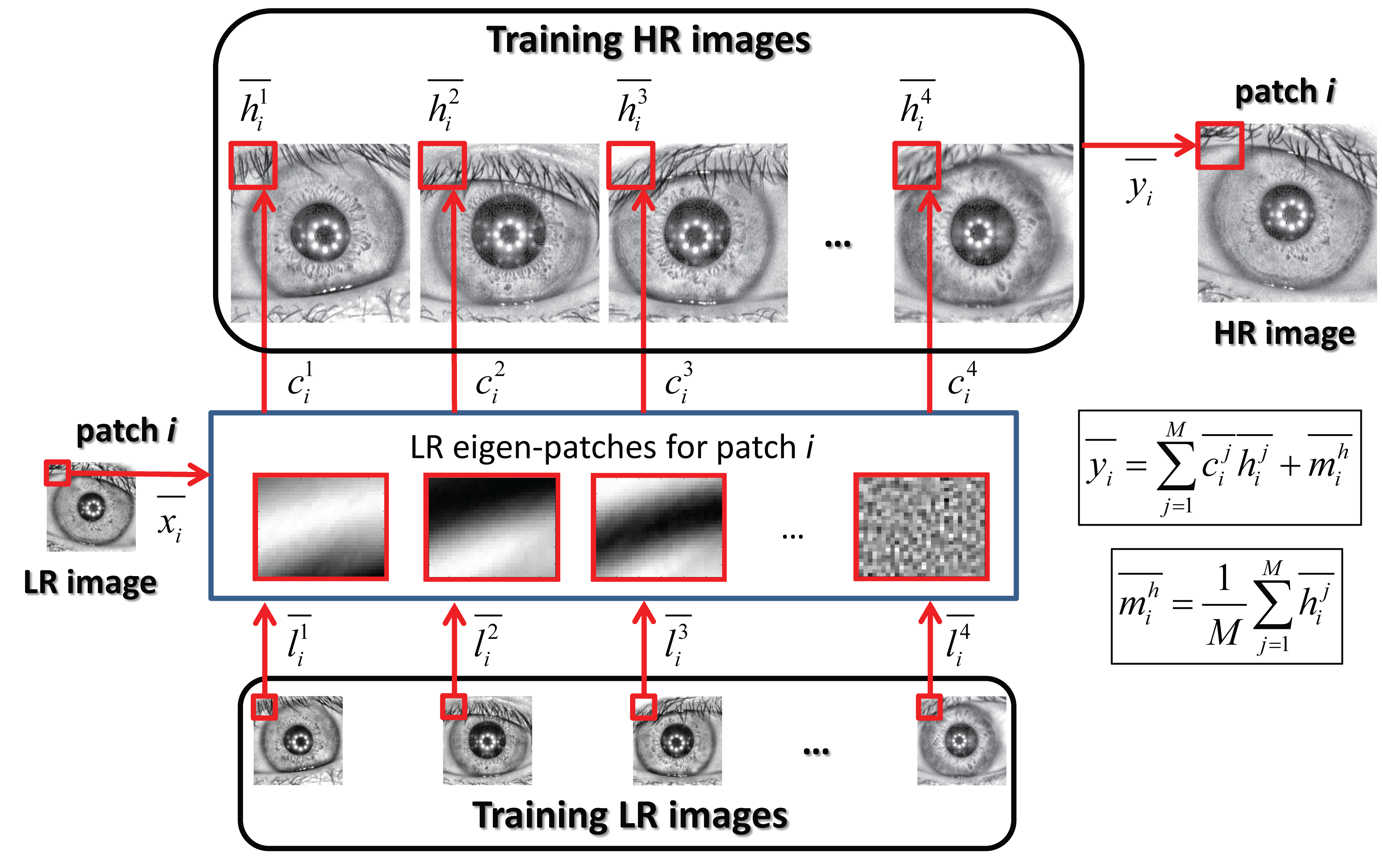}
     \caption{Eigen-patch hallucination step.}
     \label{fig:system1}
\end{figure}

\begin{figure}[htb]
     \centering
     \includegraphics[width=.3\textwidth]{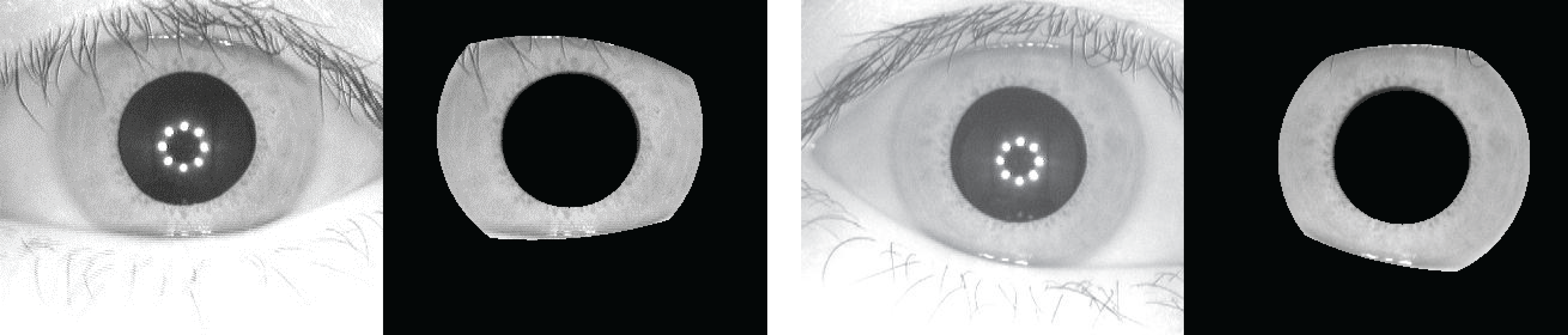}
     \caption{Example of images of the CASIA Interval v3 database
     with the annotated circles modeling iris boundaries and eyelids.}
     \label{fig:annotation}
\end{figure}

\begin{figure}[htb]
     \centering
     \includegraphics[width=.3\textwidth]{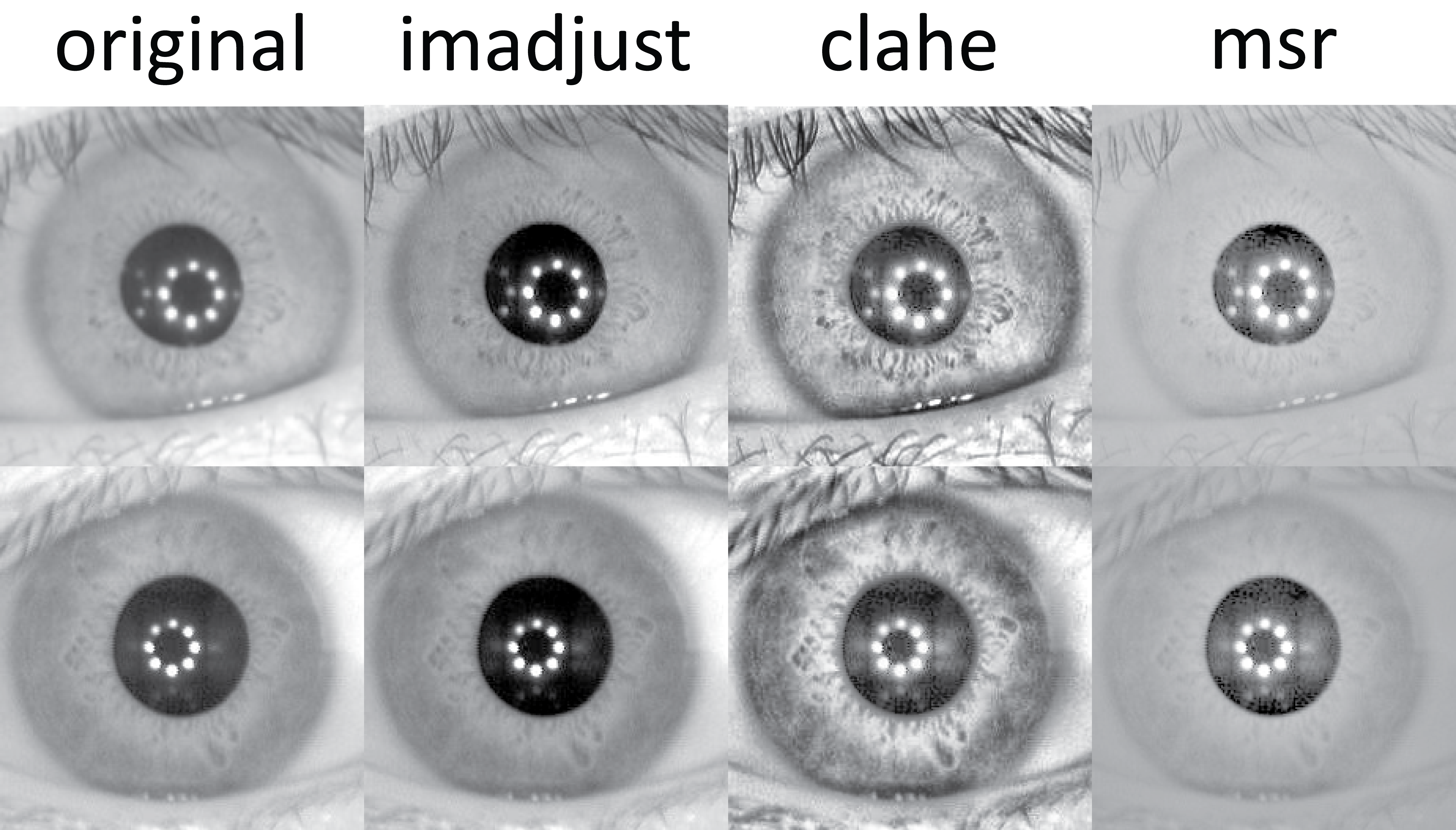}
     \caption{Example of HR original images with different contrast enhancement techniques.}
     \label{fig:images-example1}
\end{figure}

In this paper, we apply an iris super-resolution technique based on
PCA Eigen-transformation of local image patches inspired by the
system of \cite{[Chen14]} for face images.
A PCA Eigen-transformation is conducted in each patch of the input
LR image. The HR patch is then reconstructed as a linear combination
of collocated HR patches of the training database. This way, every
patch has its own optimal reconstruction coefficients, allowing to
preserve local image information. Prior to the hallucination
process, iris images are aligned with respect to the pupil center,
since alignment is critical for the performance of SR systems.
Despite the patch-based SR approach used is not new, we contribute
with its implementation to iris images, and particularly with the
application (and fusion) of two iris matchers to the reconstructed
images.
We also test different global and local contrast enhancement
techniques, previously used in iris or face studies, before carrying
out image reconstruction.
In our experiments, we employ the CASIA-IrisV3-Interval database
\cite{[CASIAdb]} of NIR iris images. We conduct verification
experiments with two iris matchers based on Log-Gabor wavelets
\cite{[MasekThesis03]} and SIFT keypoints \cite{[Lowe04]}.
Log-Gabor exploit texture information globally (across the entire
iris image), while SIFT exploit local features (in discrete key
points), therefore our motivation is to employ features that are
diverse in nature, and reveal if they behave differently.
The presented hallucination method is superior to bilinear and
bicubic interpolations, with better recognition rates as image
resolution decreases.
Regarding image enhancement, one of the matchers (SIFT) is benefited
by the use of some contrast enhancement before up-sampling, whereas
the LG matcher works better without any kind of pre-processing.
This is also exploited in the fusion of the two systems, since
better fusion results are obtained if images used with each matcher
have applied its \emph{optimum} enhancement. In our experiments,
performance of the fusion can be pushed to a EER below 5\% for any
given down-sampling factor (which in this paper includes images of
only 13$\times$13)
In addition, we observe that recognition performance is not degraded
significantly with any matcher until a down-sampling factor of 8
(image size of 29$\times$29).
This allows for example to reduce storage or data transmission
requirements, and it contributes to the feasibility of
'at-a-distance' iris acquisition as well, which is one of the most
important issues for these technologies to achieve massive adoption
\cite{[Jain10]}.

\section{Iris Hallucination Procedure}
\label{sec:method}

The structure of the hallucination method is shown in
Figure~\ref{fig:system}. It is based on the eigen-patch
hallucination method for face images of~\cite{[Chen14]}. The system
is described next.

\begin{figure*}[htb]
     \centering
     \includegraphics[width=.88\textwidth]{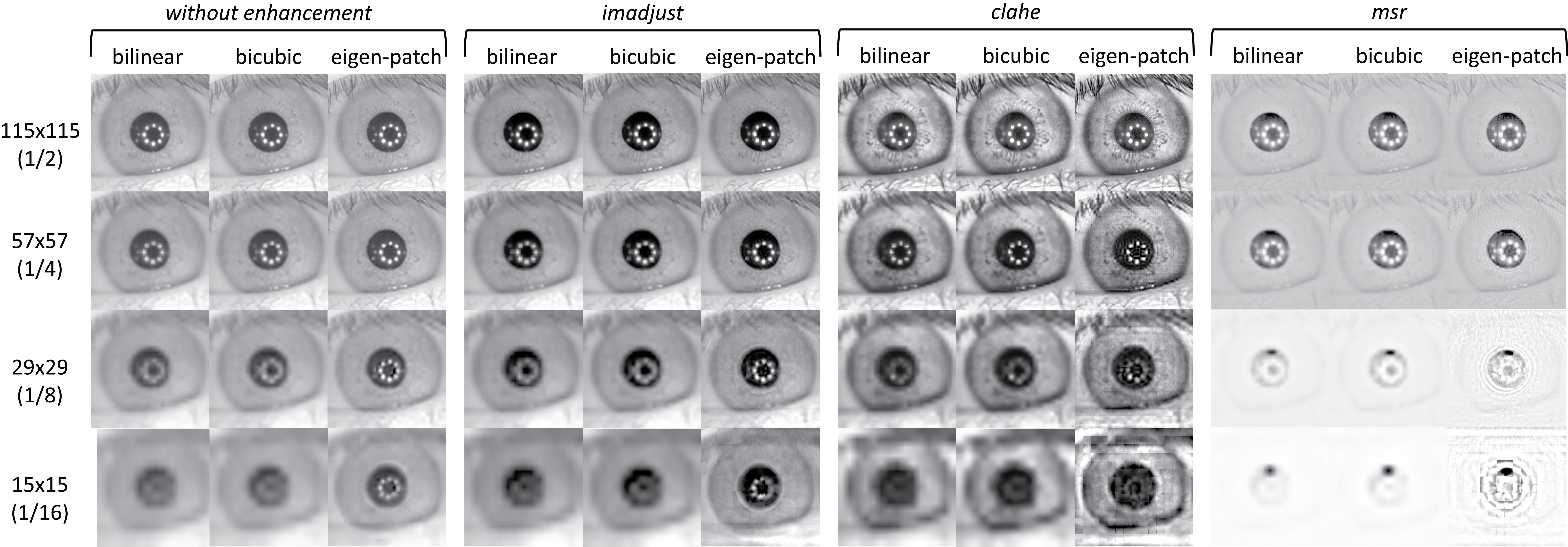}
     \caption{Resulting HR hallucinated images for different down-sampling factors.
            The original HR images are shown in Figure~\ref{fig:images-example1} (first row).}
     \label{fig:images-example}
\end{figure*}

\subsection{Eigen-Patch Hallucination}

Given an input low resolution (LR) iris image $\overline X$, it is
first preprocessed to enhance contrast; then, it is separated into
$N$ overlapping patches
$\left\{ {\overline x } \right\} = \left\{ {\overline {x_1 }
,\overline {x_2 } , \cdots ,\overline {x_N } } \right\} $.
Two super sets of basis patches are computed for each LR patch
$\overline x _i$, separate from the input image $\overline X$, from
collocated patches of a training database of high resolution images
$\{\bar H\}$. One of the super sets, $\left\{ {\overline {h_i^1 }
,\overline {h_i^2 } , \cdots ,\overline {h_i^M } } \right\} $, is
obtained
from collocated (HR) patches. By degradation (low-pass filtering and
down-sampling) a low resolution database $\{\bar L\}$ is obtained
from  $\{\bar H\}$, and the other super set, $\left\{ {\overline
{l_i^1 } ,\overline {l_i^2 } , \cdots ,\overline {l_i^M } } \right\}
$,  is obtained similarly, but for $\{\bar L\}$. $M$ is the size
(number of images) of the training set.
The same contrast enhancement procedure is applied to the HR iris
images of the training set before down-sampling.
A PCA Eigen-transformation is then conducted in each input LR patch
$\overline x _i$ using the collocated patches $\left\{ {\overline
{l_i^1 } ,\overline {l_i^2 } , \cdots ,\overline {l_i^M } } \right\}
$ of the LR training images to obtain the optimal reconstruction
weights $\overline {c_i }  = \left\{ {c_i^1 ,c_i^2 , \cdots ,c_i^M }
\right\}$ of each patch (see Figure~\ref{fig:system1}). By allowing
each LR patch of the input image to have its own optimal
reconstruction weights, the HR patch will be closer to the input LR
patch, therefore more local information can be preserved and less
reconstruction artifacts appear. Once the reconstruction weights
$\overline {c_i }$ of each patch are obtained, the HR patches are
rendered using the collocated patches of the HR images of the
training set $\left\{ {\overline H } \right\}$. The reconstruction
coefficients of the input image $\overline X$ using the LR patches
is carried on to weight the HR basis set, which yields the
preliminary reconstructed HR iris image $\overline Y '$, after
averaging the overlapping regions.
Additional details of this Eigen-transformation procedure can be
obtained in \cite{[Chen14]}.

\subsection{Image Re-projection}

A re-projection step is applied to $\overline Y '$ to reduce
artifacts and make the output image $\overline Y$ more similar to
the input LR image $\overline X$. The image $\overline Y '$ is
re-projected to $\overline X$ via $
\overline Y ^{t + 1}  = \overline Y ^t  - \tau U\left( {B\left(
{DB\overline Y ^t  - \overline X } \right)} \right)
$ where $U$ is the up-sampling matrix. The process stops when
$|\overline Y ^{t + 1}-\overline Y ^{t}|<\epsilon$. Here we use
$\tau$=0.02 and $\epsilon=10^{ - 5}$.

\section{Iris Matchers}
\label{sec:matcher}

We conduct matching experiments of iris images using two different
systems based on 1D Log-Gabor filters (LG) \cite{[MasekThesis03]}
and
the SIFT operator \cite{[Lowe04]}.
In LG, the iris region is first unwrapped to a normalized rectangle
of 20$\times$240 pixels using the Daugman's rubber sheet model
\cite{[Daugman04]} and next, a 1D Log-Gabor wavelet is applied plus
phase binary quantization to 4 levels. Matching between binary
vectors is done using the normalized Hamming distance
\cite{[Daugman04]}, which incorporates noise mask, so only
significant bits are used. Rotation is accounted for by shifting the
grid of the query image in counter- and clock-wise directions, and
selecting the lowest distance, which corresponds to the best match
between two templates.
In the SIFT matcher, SIFT key points are directly extracted from the
iris region (without unwrapping), and the recognition metric is the
number of matched key points, normalized by the average number of
detected keypoints in the two images under comparison.
The LG implementation is from Libor Masek code
\cite{[MasekThesis03]}, using its default parameters (optimized for
CASIA images, which we employ here as well).
%
The SIFT method uses a free toolkit for feature extraction and
matching\footnote{http://vision.ucla.edu/~vedaldi/code/sift/assets/sift/index.html},
with the adaptations described in \cite{[Alonso09]} (particularly,
it includes a post-processing step to remove spurious matching
points using geometric constraints).
The iris region and corresponding noise mask for feature extraction
and matching is obtained by manual annotation of the database used,
as shown in Figure~\ref{fig:annotation} (more information is
provided in the experimental setup).

\section{Experimental Framework}
\label{sec:framework}

We use the CASIA Interval v3 iris database \cite{[CASIAdb]} for our
experiments. It has 2,655 NIR images of 280$\times$320 pixels from
249 contributors captured in 2 sessions with a close-up iris camera,
totalling 396 different eyes (the number of images per contributor
and per session is not constant). Manual annotation of this database
is available \cite{[Alonso15],[Hofbauer14]}, which has been used as
input for our experiments. All images of the database are resized
via bicubic interpolation to have the same sclera radius (we choose
as target radius the average sclera radius $R$=105 of the whole
database, given by the groundtruth). Then, images are aligned by
extracting a square region of 231$\times$231 around the pupil center
(corresponding to about 1.1$\times$$R$). In case that such
extraction is not possible (for example if the eye is close to an
image side), the image is discarded. After this procedure, 1,872
images remain, which will be used for our experiments.

\begin{figure*}[htb]
     \centering
     \includegraphics[width=.88\textwidth]{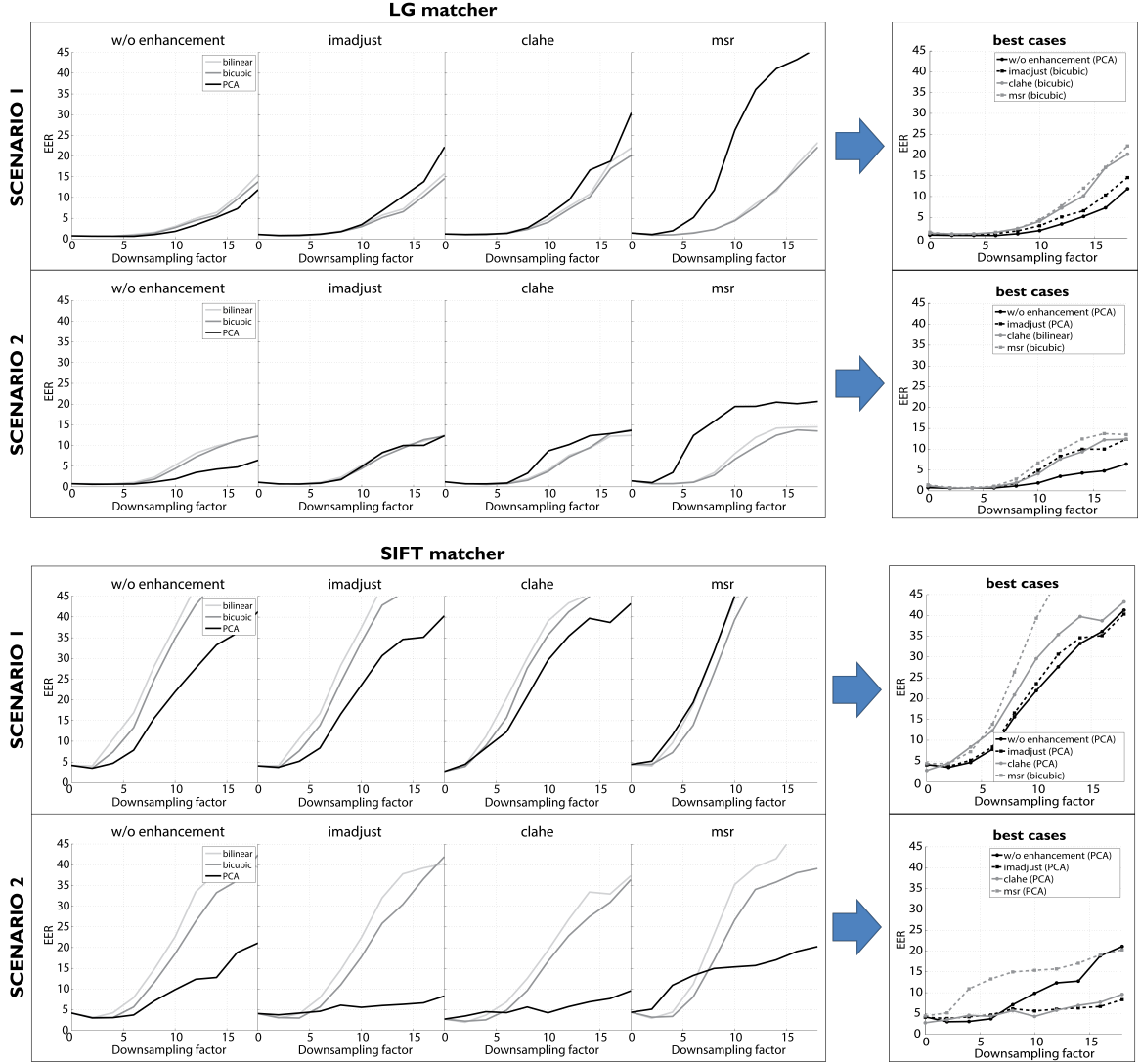}
     \caption{Verification results (EER) of the two scenarios
     considered with different contrast enhancement techniques.}
     \label{fig:EER-matchers}
\end{figure*}

The dataset of aligned images has been divided into two sets, a
training set comprised of images from the first 116 users ($M$=925
images) used to train the eigen-patch hallucination method, and a
test set comprised of the remaining 133 users (947 images) which is
used for validation. We perform verification experiments with the
iris matchers in the test set. We consider each eye as a different
user. Genuine matches are obtained by comparing each image of a user
to the remaining images of the same user, avoiding symmetric
matches. Impostor matches are obtained by comparing the $1^{st}$
image of a user to the $2^{nd}$ image of the remaining users. With
this procedure, we obtain 2,607 genuine and 19,537 impostor scores.

\section{Results}
\label{sec:results}

The 947 iris images of the test set are used as our high resolution
(HR) reference images. We then down-sample these images via bicubic
interpolation by a factor of $2n$ (i.e. the image is resized to
$1/(2n)$ of the original HR size), and the down-sampled images are
used as input LR images, from which hallucinated HR images are
extracted.
This simulated down-sampling is the approach followed in most of the
previous super-resolution studies \cite{[Wang14]}, mainly due to the
lack of databases with low-resolution and corresponding
high-resolution reference images.
We test until a down-sampling factor of 18 (corresponding to a LR
image size of 13$\times$13).
We also extract the normalized iris region (size 20$\times$240) from
both the hallucinated HR and the reference HR images, as well as LG
and SIFT features, according to the algorithms of
Section~\ref{sec:matcher}.
%
%
We compare our method with bicubic and bilinear interpolation as
well. Figure~\ref{fig:images-example} shows the hallucinated images
(only for a selection of down-sampling factors for the sake of
space).

\begin{figure}[htb]
     \centering
     \includegraphics[width=.38\textwidth]{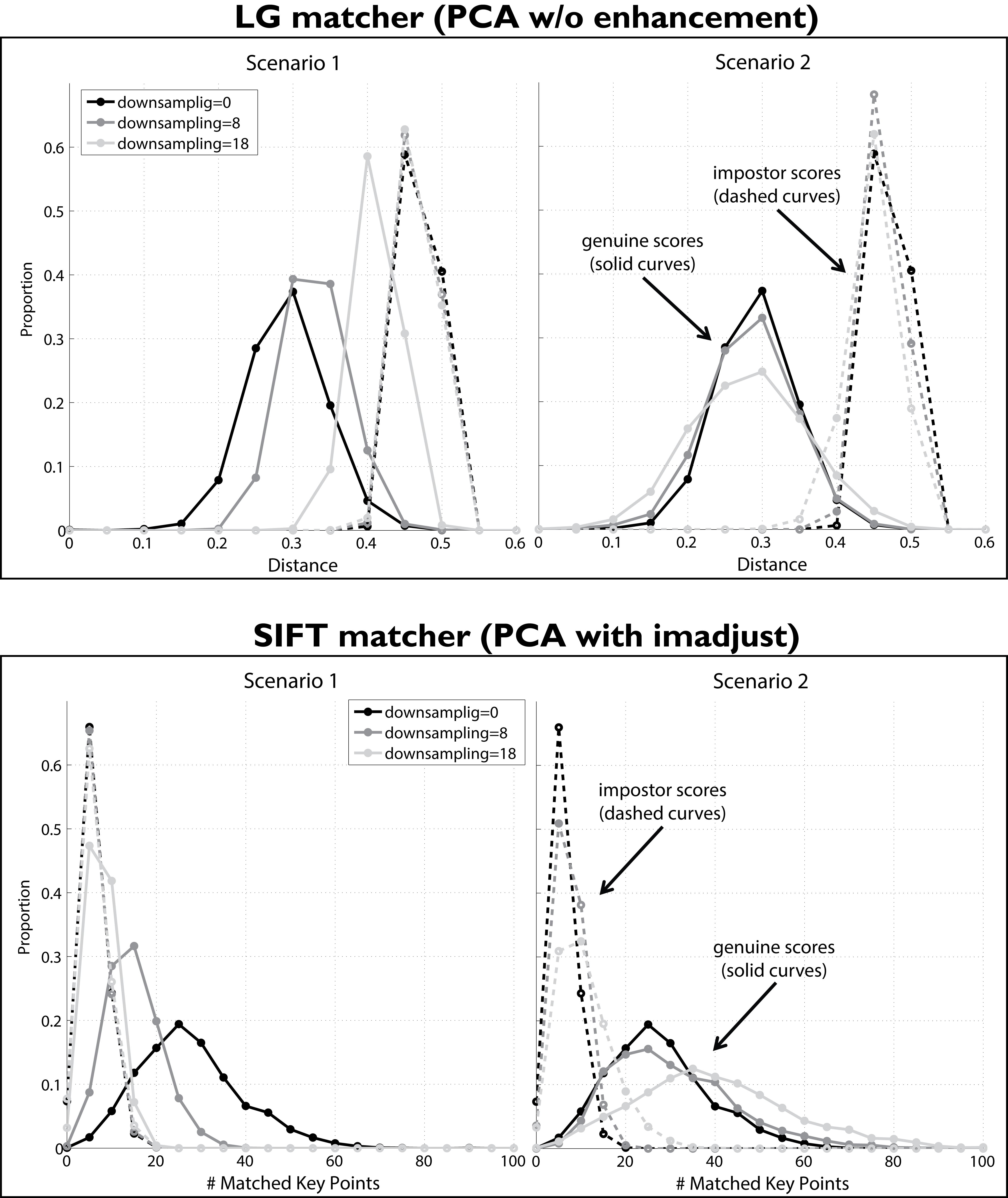}
     \caption{Score distribution of the LG and SIFT matchers for the two
     scenarios considered (selection of best cases of Figure~\ref{fig:EER-matchers}).
     Results are given for selected down-sampling factors. Solid curves correspond to genuine
     scores distributions, whereas dashed curves correspond to impostor scores distributions.}
     \label{fig:score-distributions}
\end{figure}

We have tested three different contrast enhancement algorithms with
iris images (see Figure~\ref{fig:images-example1}):
mapping of intensity values such that 1\% of data is saturated at
low and high intensities (Matlab command \emph{imadjust}),
Contrast-Limited Adaptive Histogram Equalization (\emph{clahe}, with
Matlab command \emph{adapthisteq}) \cite{[Zuiderveld94clahe]}, and
Multi-Scale Retinex (\emph{msr}) \cite{[Jobson97msr]}.
We employ the Multi-Scale Retinex implementation of the INFace
(Illumination Normalization techniques for robust Face recognition)
toolbox v2.1 \cite{[Struc11],[Struc09]}.
\emph{Imadjust} operates over the whole image, whereas \emph{clahe}
and \emph{msr} operate on local image regions or scales. The reasons
for our choices is that \emph{clahe} is usually preferred in iris
studies over other enhancement methods \cite{[Rathgeb10]}, whereas
\emph{msr} has been demonstrated to have superior performance on
face images \cite{[Juefei-Xu14]}. In our experiments with
\emph{clahe}, the number of tiles with HR images are 8$\times$8
(rows$\times$columns), which are proportionally adjusted for LR
images. With \emph{msr}, sizes of the Gaussian smoothing filters are
set to 13, 27 and 37 for HR images, which are adjusted
proportionally for LR images.

The performance of the hallucination algorithm is measured by
reporting verification experiments using hallucinated HR images.
We do not report other reconstruction measures traditionally used in
super-resolution literature (e.g. PSNR) since the aim of applying SR
algorithms in biometrics is enhancing recognition performance
\cite{[Nguyen12]}. As will be observed here, performance of the two
matchers employed is affected in a different way, so even if the
reconstruction is `good' (as measured by a general-scene indicator),
performance of a matcher may not follow. Therefore, we believe that
focusing on recognition performance is of higher interest.
Two scenarios are considered: $1$) enrolment samples taken from
original HR input images, and query samples from hallucinated HR
images; and $2$) both enrolment and query samples taken from
hallucinated HR images. The first case simulates a controlled
enrolment scenario with good quality images, while the second case
simulates a totally uncontrolled scenario (albeit for simplicity,
enrolment and query samples have similar resolution in our
experiments). Results are given in Figure~\ref{fig:EER-matchers}.
The following observations can be made from these results:

\begin{figure*}[htb]
     \centering
     \includegraphics[width=.88\textwidth]{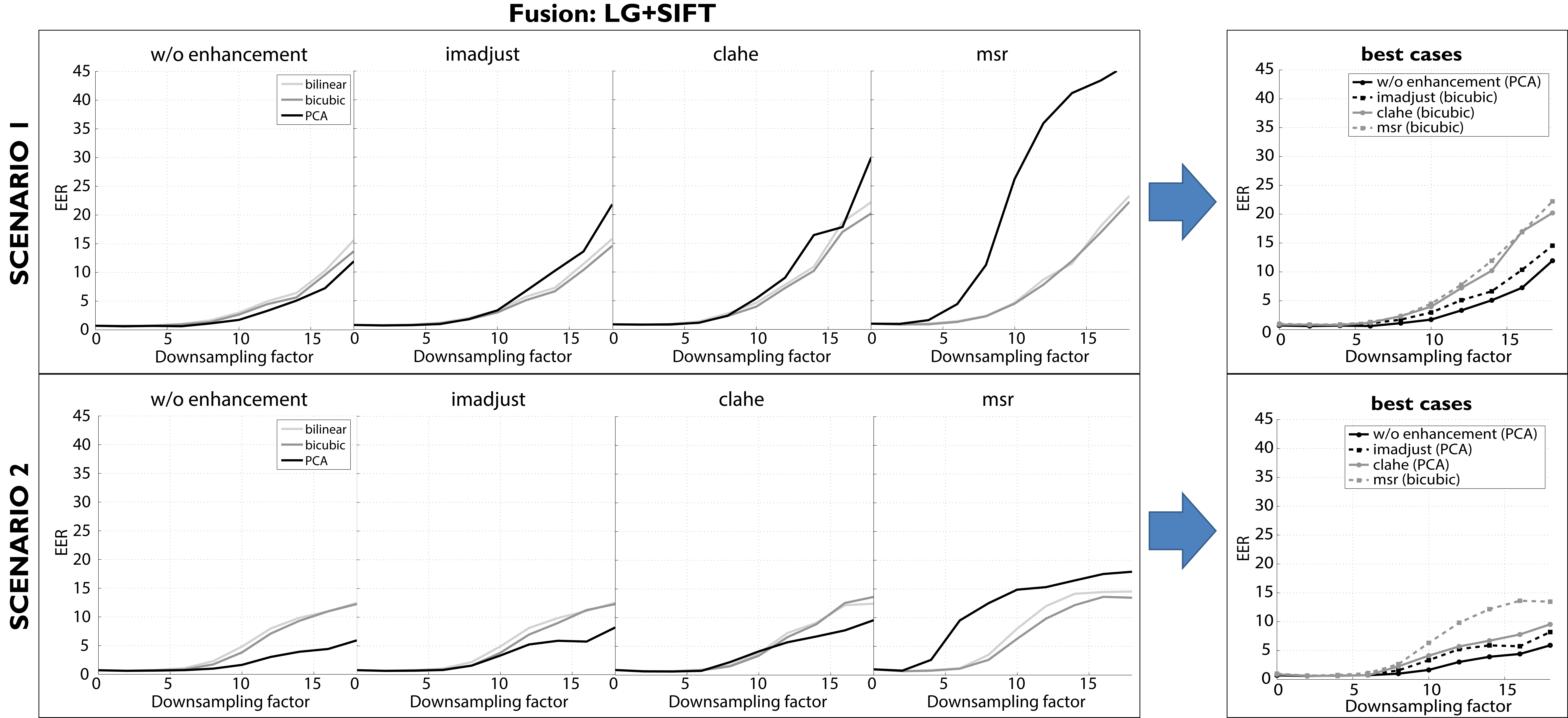}
     \caption{Fusion results (EER) of the two scenarios
     considered with different contrast enhancement techniques.}
     \label{fig:EER-matchers-fusion}
\end{figure*}

\begin{figure*}[htb]
     \centering
     \includegraphics[width=.88\textwidth]{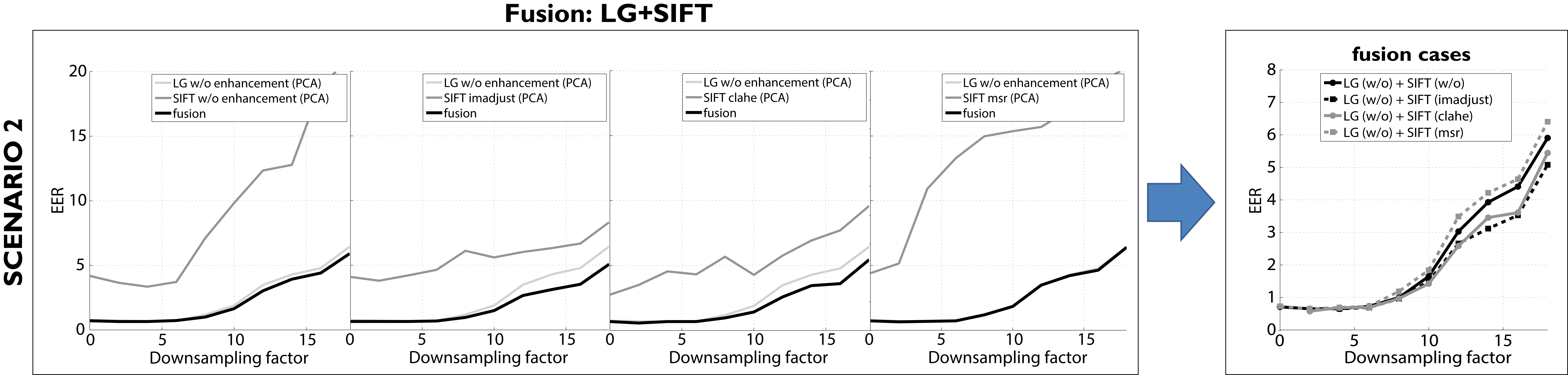}
     \caption{Fusion results (EER) of the LG matcher (without
     enhancement) and the SIFT matcher (with different enhancements).
     Results are given for scenario 2 and eigen-patch (PCA) reconstruction only.}
     \label{fig:EER-matchers-fusion1}
\end{figure*}

\begin{figure*}[htb]
     \centering
     \includegraphics[width=.88\textwidth]{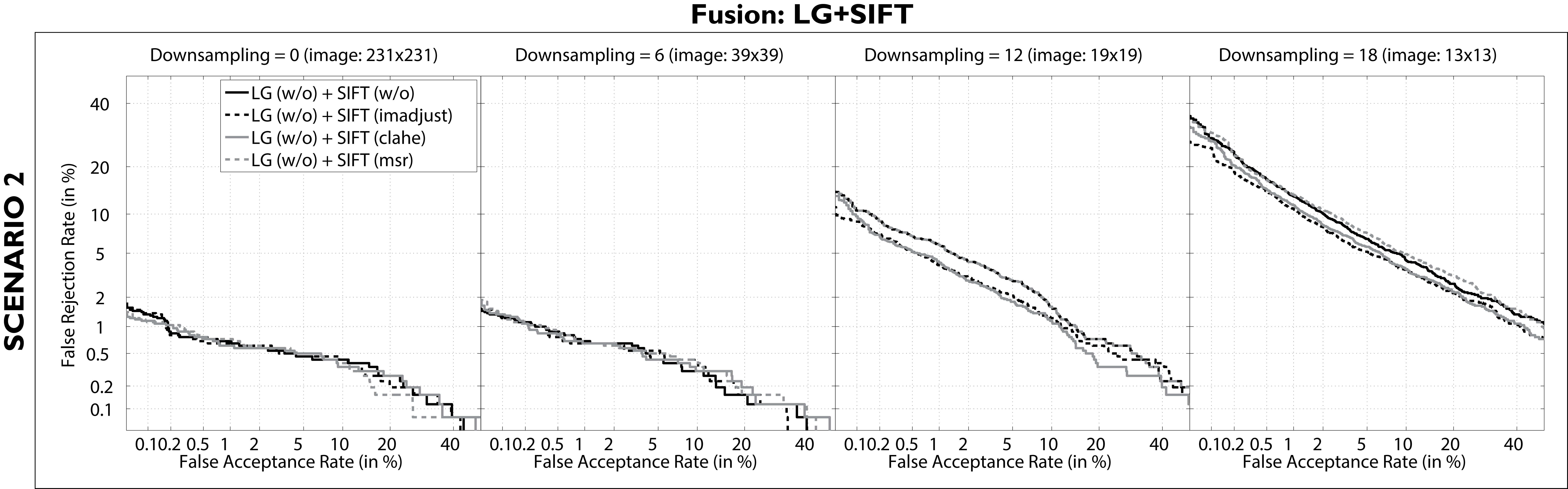}
     \caption{Fusion results (DET curves) of the LG matcher (without
     enhancement) and the SIFT matcher (with different enhancements) for several selected down-sampling factors.
     Results are given for scenario 2 and eigen-patch (PCA) reconstruction only.}
     \label{fig:EER-matchers-fusion2}
\end{figure*}

\begin{itemize}
    \item The performance of the proposed hallucination method is
    very similar to bilinear and bicubic interpolations for small
    down-sampling factors, but at low resolutions, the best
    (smallest) error rates are obtained in general with
    the proposed eigen-patch hallucination method for both matchers
    (observe the last column of `best cases' in
    Figure~\ref{fig:EER-matchers}).
    In addition, when comparing the two matchers, the best absolute
    performance is given by the LG matcher.

    \item The performance of both scenarios is similar up to a
    certain down-sampling factor. However, at low resolutions,
    the performance of scenario 2 is much better than scenario 1, and
    this phenomenon is observed with both matchers.
    We analyze
    this effect further by plotting in
    Figure~\ref{fig:score-distributions} the score distributions of
    the two matchers for the best cases of
    Figure~\ref{fig:EER-matchers}
    (for consistency, results with the
    SIFT matcher are reported for both scenarios using \emph{imadjust}
    enhancement).
    In scenario 1, the distribution of genuine scores
    is shifted towards the impostor distribution for both matchers
    as resolution decreases. This means that the
    `similarity' between hallucinated HR images and original HR
    images is reduced as the
    size of input LR images is reduced. In other words, the
    information recovered by the reconstruction algorithm does not
    fully resemble the information found in the original HR image
    (at least measured by the features employed).
    %
    %
    In scenario 2, the distribution of genuine scores is not
    significantly shifted as resolution changes, but it tends
    to be more spread instead.
    In this case, both gallery and probe images undergo the same
    down-sampling and up-sampling procedure, therefore they keep
    more `similarity' in average.
    A collateral effect of this phenomena however is that the distribution of
    impostor scores shifts towards the genuine distribution for
    extreme down-sampling, meaning that hallucinated HR images
    of different users tend to be more `similar' (under the features
    used here) as they undergo the mentioned down-sampling and up-sampling
    procedure.

    \item The difference in performance between scenario 1 and
    scenario 2 at low resolutions is much more evident with the SIFT
    matcher: EER in scenario 1 goes above 40\% for a down-sampling
    factor of 18, whereas best EERs in scenario 2 are kept below
    10\%.
    With the LG matcher, the best EER in scenario 1 is below $\sim$12\%,
    whereas for scenario 2 is below $\sim$6.5\%. This suggests that the
    image properties captured by this matcher (using a Log-Gabor wavelet)
    are less sensitive to changes due to extreme down-sampling. Or
    seen in a different way, the eigen-patch reconstruction algorithm employed
    is better in recovering the image properties analyzed by the LG
    matcher than the image properties analyzed by the SIFT matcher.
    However, the SIFT matcher is more resilient to
    reductions in image resolution when operating in scenario 2
    (see bottom left plot of Figure~\ref{fig:EER-matchers}),
    with the EER going from 3-4\%
    with no down-sampling to only $\sim$8\% with a down-sampling factor of 18.
    On the other hand, the LG matcher goes from 0.76\% to
    6.44\%, thus increasing EER by an order of magnitude
    (second plot, last column of Figure~\ref{fig:EER-matchers})

    \item From the last column of `best cases' in Figure~\ref{fig:EER-matchers},
    it can be observed that performance in scenario 2 does not start to
    degrade significantly until a down-sampling factor of 8 (image
    size of 29$\times$29). This is true for the two matchers
    employed, suggesting that the size of both gallery and probe
    images can be kept low without sacrificing performance. This has
    positive implications in terms of low storage or low data
    transmission needs (original image size is 231$\times$231). With
    the LG matcher, bilinear and bicubic interpolations show similar
    performance than the proposed method for small down-sampling
    factors; however, with the SIFT matcher, bilinear and bicubic
    interpolations degrade more rapidly than the PCA eigen-patch
    reconstruction, thus highlighting the benefits of the proposed
    method even with small down-sampling.
    It is also remarkable the relatively low EER figures obtained for the most
    extreme case of down-sampling analyzed here (a factor of 18,
    which corresponds to an image size of just 13$\times$13):
    EER of $\sim$6\% with the LG matcher and
    $\sim$8\% with the SIFT matcher.

    \item With regards to the contrast enhancement algorithms employed,
    they do not benefit the two matchers equally.
    The best performance with the LG matcher at low resolutions
    is obtained without any kind of enhancement, but the SIFT
    matcher is benefited of the use of \emph{imadjust} or
    \emph{clahe} algorithms.
    It is of relevance as well that some enhancement techniques
    produces that bilinear or bicubic interpolations work better
    than eigen-patch interpolation
    with the LG matcher (e.g. see \emph{clahe}, which is a popular
    enhancement algorithm in iris recognition research \cite{[Rathgeb10]}).
    There is also consensus in our results on the fact that \emph{msr} produces the
    worst results with both matchers.
    It is also worth noting that \emph{imadjust} (which operates over the
    whole image) and \emph{clahe} (which operates on local patches)
    result in similar performance with both matchers in scenario 2,
    see last column of Figure~\ref{fig:EER-matchers}.

    \item Although it is not the scope of this paper, it can be seen
    in Figure~\ref{fig:EER-matchers} that some curves show a slight
    improvement in performance after a small down-sampling of just
    2 or 4. This suggests that the two matchers used here may
    benefit of a small image smoothing when using images at the
    original (high) resolution.

\end{itemize}

We then carry out fusion experiments using linear logistic
regression. Given $N$ matchers ($N$=2 in our case) which output the
scores ($s_{1j}, s_{2j}, ... s_{Nj}$) for an input trial $j$, a
linear fusion is: $f_j = a_0 + a_1 \cdot s_{1j} + a_2 \cdot s_{2j} +
... + a_N \cdot s_{Nj}$. The weights $a_{0}, a_{1}, ... a_{N}$ are
trained via logistic regression as described in \cite{[Alonso08]}.
We use this trained fusion approach because it has shown better
performance than simple fusion rules (like the mean or the sum rule)
in previous works, as in the one reported above.
Results are given in Figure~\ref{fig:EER-matchers-fusion}.
Experiments reported in this Figure make use of the same contrast
enhancement technique with the two matchers. For this reason,
despite the fact that SIFT is benefited by the use of
\emph{imadjust} or \emph{clahe}, the best fusion results are
obtained without any kind of enhancement, since LG dominates in the
fusion due to its smaller individual EER. To counteract this effect,
we test several LG+SIFT fusion combinations after applying the
different enhancement techniques to the SIFT matcher. Results of
this procedure are reported in
Figure~\ref{fig:EER-matchers-fusion1}. As it can be observed, better
results are obtained at low resolutions when \emph{imadjust} or
\emph{clahe} are applied to the SIFT matcher, obtaining higher
improvements w.r.t the best individual matcher (i.e. the LG
matcher). We also report in Figure~\ref{fig:EER-matchers-fusion2}
the DET curves of the fusion combinations of
Figure~\ref{fig:EER-matchers-fusion1} for several (selected)
down-sampling factors. Despite the fact that the EER of the
different fusion combinations has been observed to be similar for
small down-sampling factors, the leftmost plot of
Figure~\ref{fig:EER-matchers-fusion2} (no down-sampling) suggest
that at low FAR or FRR, there is benefit from applying enhancement
to the SIFT matcher too (it is even remarkable in this case that at
low FRR, the best enhancement is \emph{msr}, which has never
appeared as the best option so far). Also, as observed above, the
two rightmost plots confirm that \emph{imadjust} or \emph{clahe}
should be applied to the SIFT matcher. It is interesting, however,
that \emph{imadjust} is more beneficial at low FAR, whereas
\emph{clahe} may be preferred at low FRR.

\section{Conclusions}
\label{sec:conclusions}

The use of more relaxed acquisition conditions will push iris
recognition towards the use of low resolution imagery
\cite{[Bowyer07]}.
In this paper, we apply an iris super-resolution technique based on
PCA Eigen-transformation of local image patches \cite{[Chen14]} to
increase the resolution of near-infrared (NIR) iris images.
We also test the use of different global and local contrast
enhancement algorithms with the eigen-patch reconstruction system.
The proposed approach is compared to bilinear and bicubic
interpolations as well.
Performance of the reconstruction algorithm is measured by reporting
verification experiments with two iris matchers based on Log-Gabor
(LG) wavelets and SIFT keypoints.
We consider two operational scenarios, one where original
high-resolution images are matched against hallucinated
high-resolution images (scenario 1, or controlled enrolment), and
another scenario where only hallucinated images are used (scenario
2, or uncontrolled scenario).
Experiments show that, at low resolutions, better performance can be
obtained with the proposed eigen-patch reconstruction method w.r.t.
bilinear or bicubic interpolation.
Regarding image enhancement, the LG matcher shows better performance
without any kind of enhancement, whereas the SIFT matcher is
benefited by the use of some contrast enhancement before image
reconstruction.
We also observe that if both gallery and probe images undergo the
same down-sampling and up-sampling procedure (scenario 2), they keep
more `similarity' than if up-sampled images are matched against
original high-resolution images (scenario 1). This is reflected by a
better recognition performance of scenario 2 w.r.t. scenario 1 at
low resolutions. However, the LG matcher is less sensitive to this
effect, as shown by smaller relative differences in performance
between the two scenarios than the SIFT matcher.
It is also worth noting that performance with any matcher is not
significantly degraded until image is down-sampled by 8 or higher
factors, allowing to use images of reduced size (of importance for
example under low storage or data transmission capabilities).
We also perform fusion experiments between the two matchers, with
results showing that performance at low resolutions can be further
improved, with EER pushed to below 5\% for any given down-sampling
factor (which in our experiments includes iris images of up to a
down-sampling factor of 18, or an image size of only 13$\times$13).

The proposed PCA method assumes linearity in finding the
reconstruction weights of each patch; in addition, it employs all
available collocated patches of the training database. Allowing
non-linearity and deriving an optimal subset of collocated patches
from the training set are two avenues that are demonstrating
superiority in preserving texture in face super-resolution studies
\cite{[Jiang14]}, which we will evaluate here too as future work.
%
We expect that these approaches will better cope with artifacts
appearing at low resolutions due to division of the image in patches
(observe bottom of Figure~\ref{fig:images-example}).
The latter could be also achieved by exploring other methods to
combine overlapping patches during reconstruction. We follow here
the predominant approach of averaging patches, which acts as a
denoising that reduces texture details and may result in
over-smoothing \cite{[Wang14]}, but other approaches should be
explored as well.
We are also aware of the limitation of comparing our system against
simple bilinear and bicubic interpolation approaches. However, as
mentioned in the introduction, research in iris super-resolution is
limited. Implementation of the improvements mentioned above will
lead to the availability of other techniques that will provide a
more comprehensive experimental framework. In addition, we have not
considered iris reconstruction-based approaches, since they demand
more than one image for reconstruction.
Another direction is increasing the number of systems involved in
the fusion, as well as the impact of using a smaller training set.
We also plan to analyze sensitivity of the proposed method to
misalignment of the eye, since alignment is very critical to proper
hallucination output.

\footnotesize

\section*{Acknowledgements}
This work was initiated while F. A.-F. was a visiting researcher at
the University of Malta, funded by EU COST Action IC1106. Author F.
A.-F. also thanks the Swedish Research Council for funding his
research, and the CAISR program of the Swedish Knowledge Foundation.

\bibliographystyle{ieee}

\end{document}